\documentclass[journal]{IAENGtran}
\usepackage{wrapfig}
\usepackage{soul}
\usepackage{multirow}
\usepackage{booktabs}
\usepackage{adjustbox}
\usepackage{graphicx}
\usepackage{algorithm}
\usepackage[noend]{algorithmic}
\usepackage{xcolor,colortbl}
\usepackage{color}
\usepackage{textcomp}
\usepackage{amsmath}

\usepackage[numbers,sort&compress,square]{natbib}
\definecolor{Gray}{gray}{0.85}
\definecolor{LightCyan}{rgb}{0.88,1,1}

\usepackage{epstopdf}
\usepackage{subfloat}
\usepackage{diagbox}
\usepackage[misc]{ifsym}
\usepackage{float}
\usepackage{lipsum}
\usepackage[font=small,labelfont=bf]{caption}
\usepackage[caption = false]{subfig}
\usepackage{caption}
\let\labelindent\relax
\usepackage{enumitem}
\setlist[itemize]{leftmargin=*}
\renewcommand{\arraystretch}{1.2}
\usepackage{lipsum} 
\usepackage{fancyhdr}
\begin{document}\sloppy

\title{
Toward Personalized Training and Skill Assessment in Robotic Minimally Invasive Surgery
}

\author{Mahtab~J.~Fard, Sattar~Ameri, and~R.~Darin~Ellis
\thanks{Manuscript received July 23, 2016; revised August 8, 2016.}
\thanks{M. J. Fard, S. Ameri and R.D. Ellis are with the Department
of Industrial and System Engineering, Wayne State University, Detroit, MI, 48202 USA e-mail: {\tt\small {fard@wayne.edu.}}}
}

\maketitle

\begin{abstract}
Despite the immense technology advancement in the surgeries the criteria of assessing the surgical skills still remains based on subjective standards. With the advent of robotic-assisted minimally invasive surgery ({\em{RMIS}}), new opportunities for objective and autonomous skill assessment is introduced. Previous works in this area are mostly based on structured-based method such as Hidden Markov Model (HMM) which need enormous pre-processing. In this study, in contrast with them, we develop a new shaped-based framework for automatically skill assessment and personalized surgical training with minimum parameter tuning. 
Our work has addressed main aspects of skill evaluation; develop gesture recognition model directly on temporal kinematic signal of robotic-assisted surgery, and build automated personalized RMIS gesture training framework which . We showed that our method, with an average accuracy of 82\% for suturing, 70\% for needle passing and 85\% for knot tying, performs better or equal than the state-of-the-art methods, while simultaneously needs minimum pre-processing, parameter tuning and provides surgeons with online feedback for their performance during training.
\end{abstract}
\begin{IAENGkeywords}
Robotic Surgery, Gesture classification, Surgical skill assessment, Time series classification, Dynamic time warping.
\end{IAENGkeywords}

\thispagestyle{fancy}
\section{Introduction}
\IAENGPARstart{T}{he} hospital operating room is a challenging work environment where surgical skills have been learned there with direct supervision of expert surgeons for many years \cite{reznick1993teaching}. This procedure is very time-consuming and subjective that cause surgeon's skill evaluation be non-robust and unreliable \cite{darzi1999assessing}.
Different surgical gestures have different levels of complexity and  the skill level of surgeon varies and can be enhanced with teaching and training \cite{Kassahun2015, ellis2015task}. 
Hence, it is important to find underlying signatures of surgeon for each surgical gesture to be able to asses and evaluate the quality of the skills that were learned. The aim of this study is to build a personalized surgical training framework and skill assessment through a quantitative methods.

With the new technology innovations, such as minimally invasive surgery and more advanced, robotic minimally invasive surgery (RMIS), the need for non-subjective based skill evaluation have been arisen \cite{van2010objective}. Although these technologies introduce new challenges in skill assessment due to the steep learning curve but on the other hand, they open a new opportunities for objective and automated surgical assessment which was not available before \cite{lalys2014surgical}. 
Current systems like {\em da Vinci} (Intuitive Surgical, Sunnyvale, CA) \cite{guthart2000intuitivetm} record motion and video data, enabling development of computational models to recognize and analyze surgeon's skills and performance through data-driven approaches.
 
The key step for autonomous skill evaluation of surgeons is to develop techniques that are capable of accurately recognizing surgical gestures \cite{fard2017eee}. These can then frame the premise for creating quantitative measures of surgical skills and consequently automatically annotate those gestures that needs more training \cite{jahanbani2016computational}. A range of techniques have been developed to assess surgical skills of junior surgeon \cite{reiley2011review}. Most of the prior work extracts features from kinematic and video data and build gesture classification models using Hidden Markov Models (HMMs) based approaches \cite{rosen2001markov, reiley2009task, Zappella2013} and descriptive curve coding (DCC) \cite{Ahmidi2015}.
However, these methods are very time-consuming, interactive and subjective which result in lack of consistency, reliability and efficiency in real-time feedback \cite{schout2010validation}.

In order to address these drawbacks, one natural approach is to develop shaped-based time series classification methods directly on temporal kinematic signal, captured during surgeries \cite{fu2011review}. 
In this paper, we extend our previous work \cite{RCS:RCS1766, fard2016machine} to investigate the feasibility of building personalized gesture training and skill assessment framework.
In this framework, the similarity of two time series determines by comparing their individual Dynamic Time Warping (DTW) point values \cite{bernad1996finding}.
Dynamic Time Warping (DTW) is a well-known technique for time series classification \cite{fu2011review}. The similarity that has been derived from DTW, can be used as an input to the $k$-Nearest Neighbors algorithm ($k$NN), a popular classification method, to classify a new data based on its similarity to other sample data \cite{bhatia2010survey}.
Our work has addressed two main aspects of skill evaluation; develop gesture recognition model directly on temporal kinematic signal of robotic-assisted surgery, and build automated personalized RMIS gesture training framework which provide online augmented feedback using the model trained in classification step. 
Using the proposed framework, one can also evaluate skill between novice and expert surgeons.

\begin{figure*}[htbp]
\centering
\scalebox{0.5}{\includegraphics{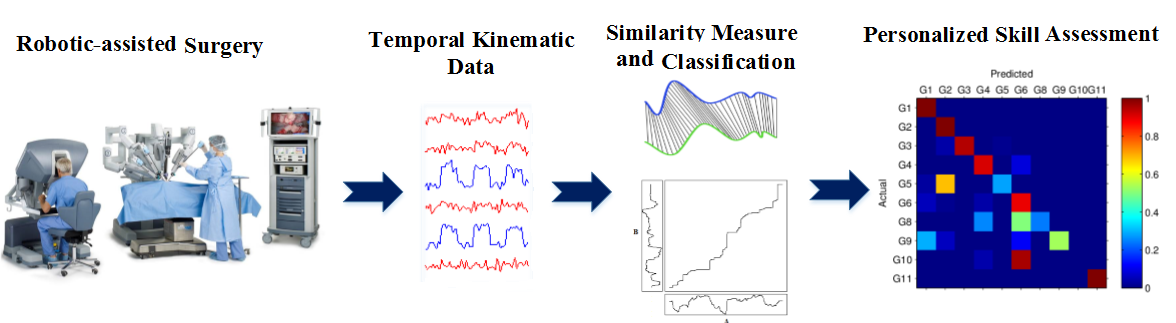}}
\caption{Personalized skill evaluation and assessment framework for robotic minimally invasive surgery.}
\label{flowchart}
\end{figure*}

\section{Method}
The aim of our work is to build a personalized framework for surgical task training. Thus, the skill evaluation framework that developed in this study, contains of two key components as shown in Fig \ref{flowchart}. The first component is to measure the similarity between surgemes performed by different surgeons and recognize them based on the $k$-Nearest Neighbor approach.
Then, based on the classification result we can evaluate the performance of surgeon for doing each gesture. Consequently, gestures that needs more training can be identified.

\subsection{Similarity Measures and Gesture Recognition}
The first component of the framework is to measure the similarity between motion time series signals of surgemes performed by surgeons. Time series data are one example for longitudinal data \cite{mahtabIEOM2015, fard2016early, 7564399}, which different method can be used to measure similarity of data.
We applied distance between two time series as a similarity measure \cite{keogh2003need}.
Shaped-based similarity measure techniques are among the well-developed methods in this area where determine the similarity of the overall shape of two time series by directly comparing their individual point values \cite{lin2012pattern}. It is in contrast with feature-based and structure-based where first features need to be extracted in order to find higher-level structures of the series. 

One of the simplest ways to estimate the distance between two time series is to use any $L_n$ norm. Given two time series A and B, their $L_n$ distance can be determined by comparing local point values as
\begin{equation}
d_{L_n}(A,B)=\bigg( \sum_{i=1}^N \left(a_i-b_i\right) ^n \bigg) ^{1/n}
\label{L-n}
\end{equation}
where $n$ is a positive integer, $N$ is the length of the time series, $a_i$ and $b_i$ are the $i^{th}$ element of time series A and B, respectively. If $n$=2 the Eq.(\ref{L-n}) defines the Euclidean distance, the most common distance measure for time series. 
Despite the simplicity and efficiency of Euclidean distance which makes it the most popular distance measure, its major drawback add a limitation to this method. It requires that both input sequences be of the same length, and it is sensitive to distortions, e.g. shifting, noise, outlier. In order to handle this problem warping distances such as Dynamic Time Warping (DTW) proposed to search for the best alignment between two time series \cite{bernad1996finding}. 

Consider ${A_p} =({a_1},{a_2},...,{a_m})$ and ${B_q} =({b_1},{b_2},...,{b_m})$ where ${A}$ and ${B}$ have $p \times m$ and $q \times m$ dimension respectively, the two sequences can be arranged as $p \times q$ matrix like the sides of a grid 
in which the distance between every possible combination of time instances ${a_i}$ and ${b_j}$ is stored. To find the best match between two sequences, a path through the grid that minimizes the overall distance is needed. This path can be efficiently found using dynamic programming. If cumulative distance $\gamma(i,j)$ as distance for current cell and the minimum of cumulative distance of adjacent elements, the distance defines as
\begin{equation}
\gamma(i,j)=d(a_i,b_j)+\min\{\gamma(i-1,j-1),\gamma(i,j-1),\gamma(i-1,j)\}
\label{dtw}
\end{equation}
where $d(a_i,b_j)$ can be calculated using Eg. (\ref{L-n}) for $n$=2.

The second component of the framework is classification algorithm based on the $k$-Nearest Neighbors ($k$NN) approach. The $k$NN is a supervised distance-based classification method. In time series classification domain, $k$-Nearest Neighbor shows promising result \cite{Chaovalitwongse2007}. Despite its simplicity, $k$-Nearest Neighbors has been very successful in classification problems \cite{bhatia2010survey}. $k$NN classifier is instance-based learning where instead of constructing a general model, it simply stores instances of training data. During the classification phase majority vote of the $k$ nearest neighbor for each point is computed. Thus, the label for the query point is assigned based on the most representatives within the nearest neighbors of the points.

\subsection{Gesture Performance Evaluation}
Once we get the classification result for each individual surgeon we are able to find the gesture that need more training or practice. For this purpose, first we need the tabulated results of gesture classifications for each surgeon into a corresponding confusion matrix (Table \ref{tab:conf-matrix}). Basically, true positives (TP) is the number of correctly classified instances and true negatives (TN) are the number of correctly classified instances that do not belong to the gesture. If a gesture is incorrectly assigned to the gesture, it is a false positive (FP) and if it is not classified as gesture instances it is a false negative (FN). 
\vspace{0.2em}
\begin{table}[!htbp]
\centering
\caption{Illustration of confusion matrix for Gesture X.}
\vspace{0.3em}
\begin{tabular}{|c|c|c|c|}
\cline{3-4} 
\multicolumn{1}{c}{} &  & \multicolumn{2}{c|}{\textbf{Predicted}}\tabularnewline
\cline{3-4} 
\multicolumn{1}{c}{} &  & Gesture X & Not Gesture X\tabularnewline
\hline 
\multirow{2}{*}{\textbf{Actual}} & Gesture X & TP & FN\tabularnewline
\cline{2-4} 
 & Not Gesture X & FP & TN\tabularnewline
\hline 
\end{tabular}
\label{tab:conf-matrix}%
\end{table}%

Based on the values in the confusion matrix, the overall accuracy for gesture classification can be defined as 
\begin{equation}
Accuracy=\frac{TP+TN}{TP+FN+FP+TN}
\label{accu}
\end{equation}
Gestures that needs more training can be identified using {\emph{recall}} which also called {\emph{sensitivity}} and is defined as
\begin{equation}
Recall=\frac{TP}{TP+FN}
\label{sen}
\end{equation}
Finally, gestures that needs more robust definition can be identified using {\emph{Precision}} 
\begin{equation}
Precision=\frac{TP}{TP+FP}
\label{sen}
\end{equation}
as it a measure of result relevancy.

\section{experimental setup}
\begin{figure}[!h]
\centering
\subfloat[Suturing]{\scalebox{.35}{\includegraphics{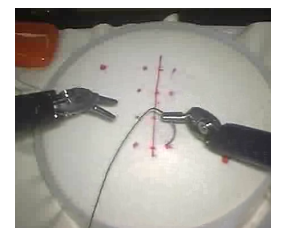}}}
\subfloat[Needle Passing]{\scalebox{.44}{\includegraphics{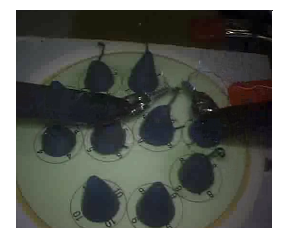}}}
\subfloat[Knot Tying]{\scalebox{.35}{\includegraphics{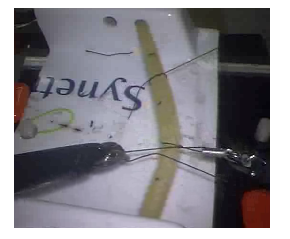}}}
\vspace{0.3em}
\caption{Three fundamental RMIS tasks \cite{gaojhu}}
\label{fig:2task}
\end{figure}

As briefly explained in introduction, we are using data presented in \cite{gaojhu}. This is comprised of data for different fundamental surgical tasks performed by eight right-handed surgeons with different skill levels (expert, intermediate and novice). Each user performed around 5 trails of the task. For each of the task, we analyze kinematic data captured using the API of the {\em da Vinci} at 30 Hz. The data includes 76 motion variables which consist of 19 features for each robotic arms, left and right master side, and the left and right patient side.
In this paper we build a personalized training framework for suturing, needle passing and knot tying (Figure \ref{fig:2task}). 

The three surgical tasks are defined as follow:
\begin{itemize}
\item {\em Suturing (SU)}: the surgeon picks up needle then proceeds to the incision and passes through tissue. Then after the needle pass, the surgeon extracts the needle out of the tissue. 
\item {\em Needle-Passing (NP)}: the surgeon picks up the needle and passes it through four small metal hoops from right to left. 
\item {\em Knot-Tying (KT)}: the surgeon picks up one end of a suture tied to flexible tube attached at its ends to the surface of the bench-top model, and ties a single loop knot. 
\end{itemize}

\vspace{-0.75em}

\begin{figure*}[ht]
\centering
\subfloat[Suturing]{\scalebox{.23}{\includegraphics[trim={.8cm 2cm 1.2cm 2cm},clip]{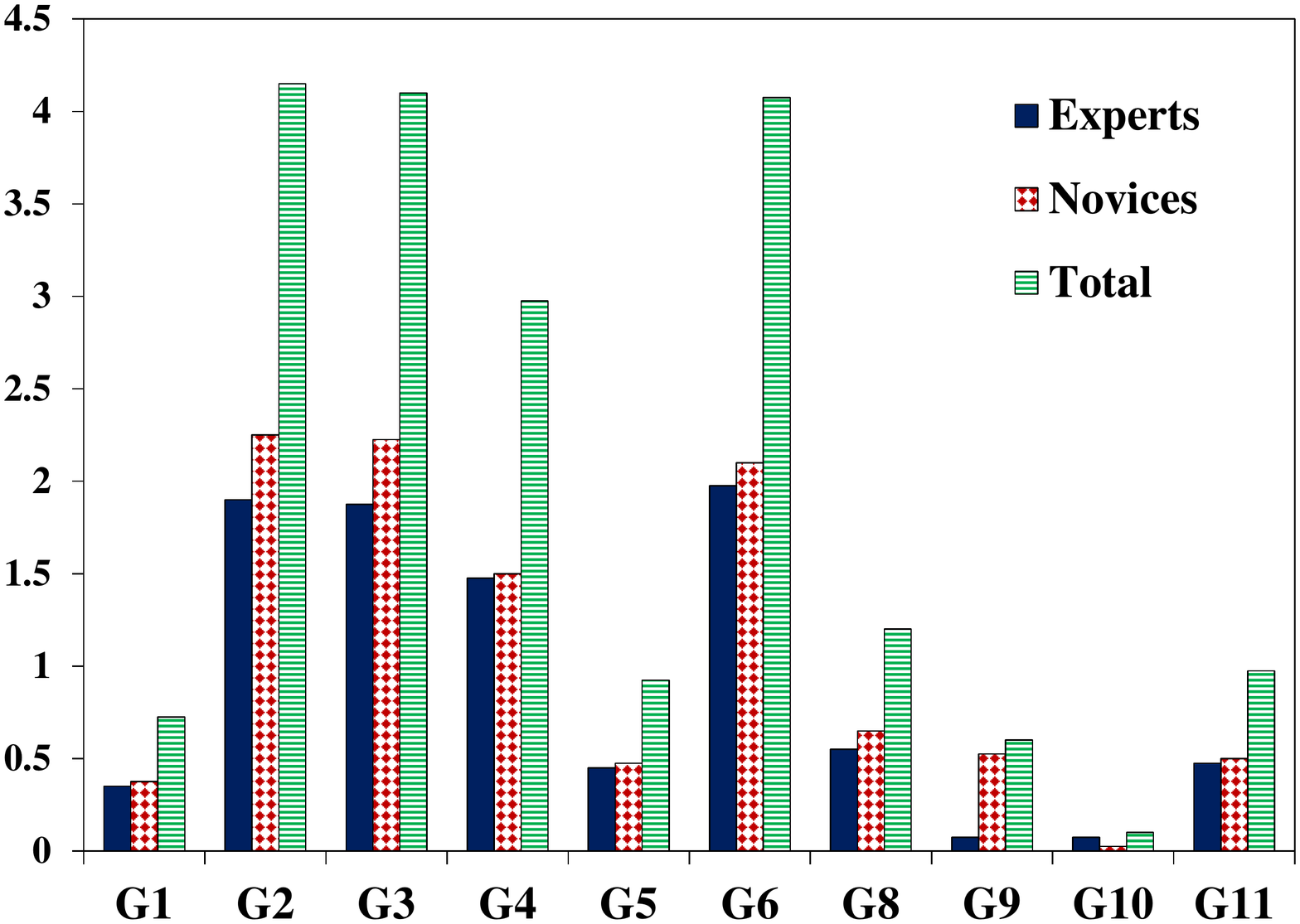}}}
\subfloat[Needle Passing]{\scalebox{.23}{\includegraphics[trim={.8cm 2cm 1.2cm 2cm},clip]{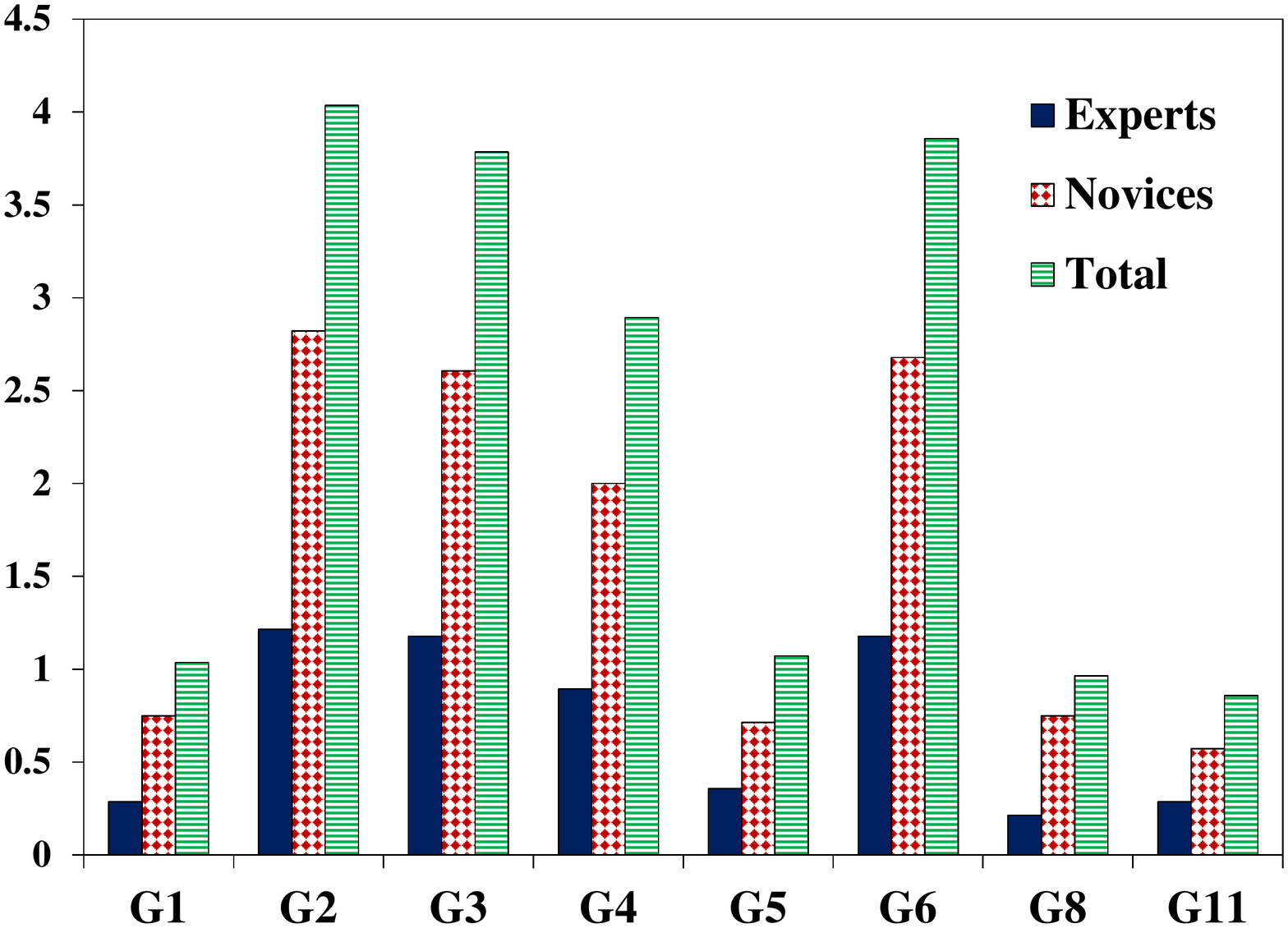}}}
\subfloat[Knot Tying]{\scalebox{.23}{\includegraphics[trim={.8cm 2cm 1.0cm 2cm},clip]{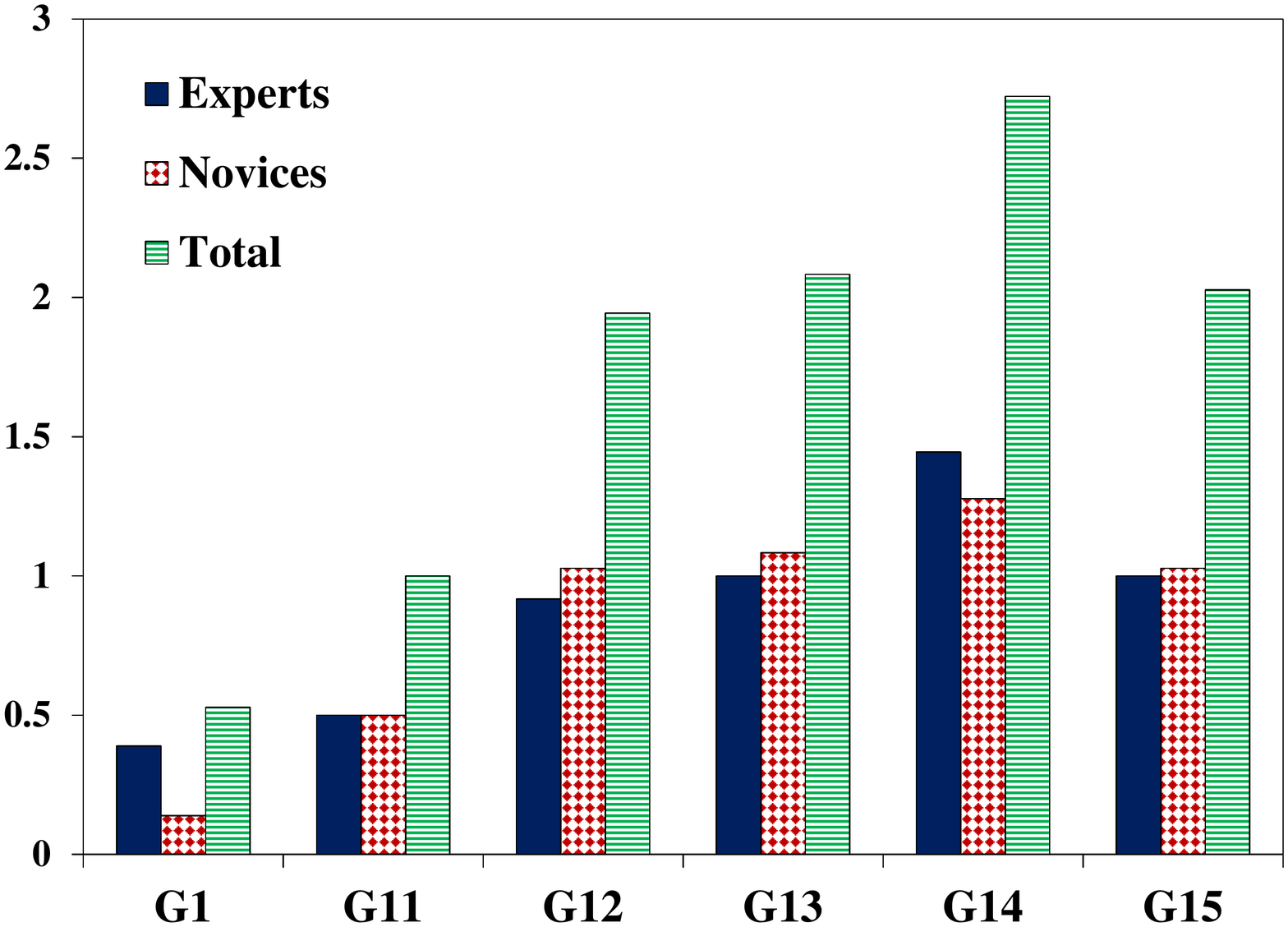}}}
\vspace{0.2em}
\caption{Frequency of surgical gestures for experts and novices during different RMIS tasks.}
\label{freq}%
\end{figure*}

The surgical activity annotation is provided manually \cite{gaojhu}. Table \ref{gesture} shows the 14 gestures that describe all these three tasks.

In order to compare the accuracy of our proposed gesture recognition framework with other methods \cite{Reiley2008, Zappella2013}, we used Leave-one-user-out (LOUO) setup in dataset. In LOUO, eight folds are created, each for one surgeon with 50 iterations. The LOUO shows the robustness of a method when a subject is not previously seen in the training data. Thus, it helps us to personalize skill assessment for each individual surgeon. 
\vspace{0.2em}

\begin{table}[htbp]
\caption{Gesture description for Suturing Needle Passing and Knot Tying}
\resizebox{\columnwidth}{!}{
\begin{tabular}{|c|l|}
\cline{2-2} 
\multicolumn{1}{c|}{} & \textbf{Gesture Index and Description}\tabularnewline
\hline 
\textbf{SU/NP/KT} & \textbf{G1: }Reaching for needle with right hand\tabularnewline
\hline 
\multirow{6}{*}{\textbf{SU/NP}} & \textbf{G2: }Positioning needle\tabularnewline
\cline{2-2} 
 & \textbf{G3: }Pushing needle through tissue\tabularnewline
\cline{2-2} 
 & \textbf{G4: }Transferring needle from left to right\tabularnewline
\cline{2-2} 
 & \textbf{G5: }Moving to center with needle in grip\tabularnewline
\cline{2-2} 
 & \textbf{G6: }Pulling suture with left hand\tabularnewline
\cline{2-2} 
& \textbf{G8: }Orienting needle\tabularnewline
\hline 
\multirow{2}{*}{\textbf{SU}} & \textbf{G9: }Using right hand to help tighten suture\tabularnewline
\cline{2-2} 
 & \textbf{G10: }Loosening more suture\tabularnewline
\hline 
\multicolumn{1}{|c|}{\textbf{SU/NP/KT}} & \textbf{G11: }Dropping suture at end and moving to end points\tabularnewline
\hline 
\multirow{4}{*}{\textbf{KT}} & \textbf{G12: }Reaching for needle with left hand\tabularnewline
\cline{2-2} 
 & \textbf{G13: }Making C loop around right hand\tabularnewline
\cline{2-2} 
 & \textbf{G14: }Reaching for suture with right hand\tabularnewline
\cline{2-2} 
 & \textbf{G15: }Pulling suture with both hands\tabularnewline
\hline 
\end{tabular}}
\label{gesture}%
\end{table}%
\vspace{-1em}

\section{Results and Discussion}
In this section, we report the experimental results of gesture recognition using the proposed classification method. 
Then, we will discuss the personalized skill evaluation framework that provides assessment to surgeons during their RMIS training.

\subsection{Distance-based Skill Evaluation}
First we start with different surgical gesture frequency analysis during RMIS tasks. Figure \ref{freq} presents the average number of surgical gesture occurrence for one surgery trial. 
It shows that, there are some surgical gestures that are very infrequent (such as G9 and G10 in suturing) and are mostly done by novice surgeons. This suggests that, those gestures are intermediate or correction positioning moves. Thus, they cannot be a good indicator when the performance of classifiers are measured. Figure \ref{freq} also indicates that the difference between number of each surgical gesture performed by novices in needle passing is significant compare to experts while for suturing and knot tying this is almost the same. Hence, one can conclude that novice surgeons might need more training for needle passing task.

\begin{figure}[htbp]
\centering
\scalebox{0.5}{\includegraphics{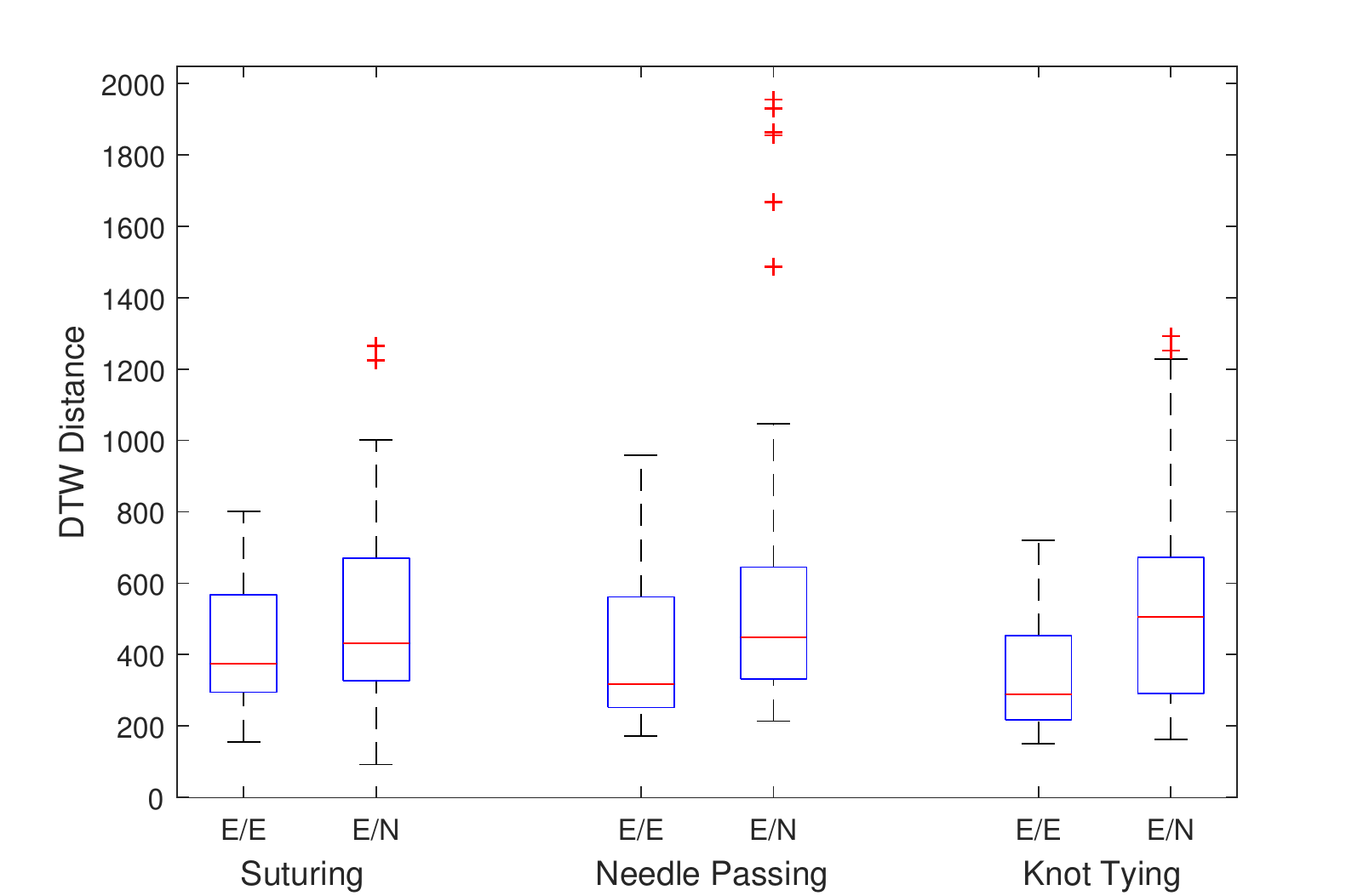}}
\vspace{0.2em}
\caption{Boxplot of DTW distance within expert surgeons (E/E) versus expert and novice (E/N) for three RMIS tasks.}
\label{fig:boxplot}
\end{figure}

\begin{figure*}[ht]
\centering
\subfloat[Suturing]{\scalebox{.23}{\includegraphics[trim={.8cm 2cm 1.4cm 2cm},clip]{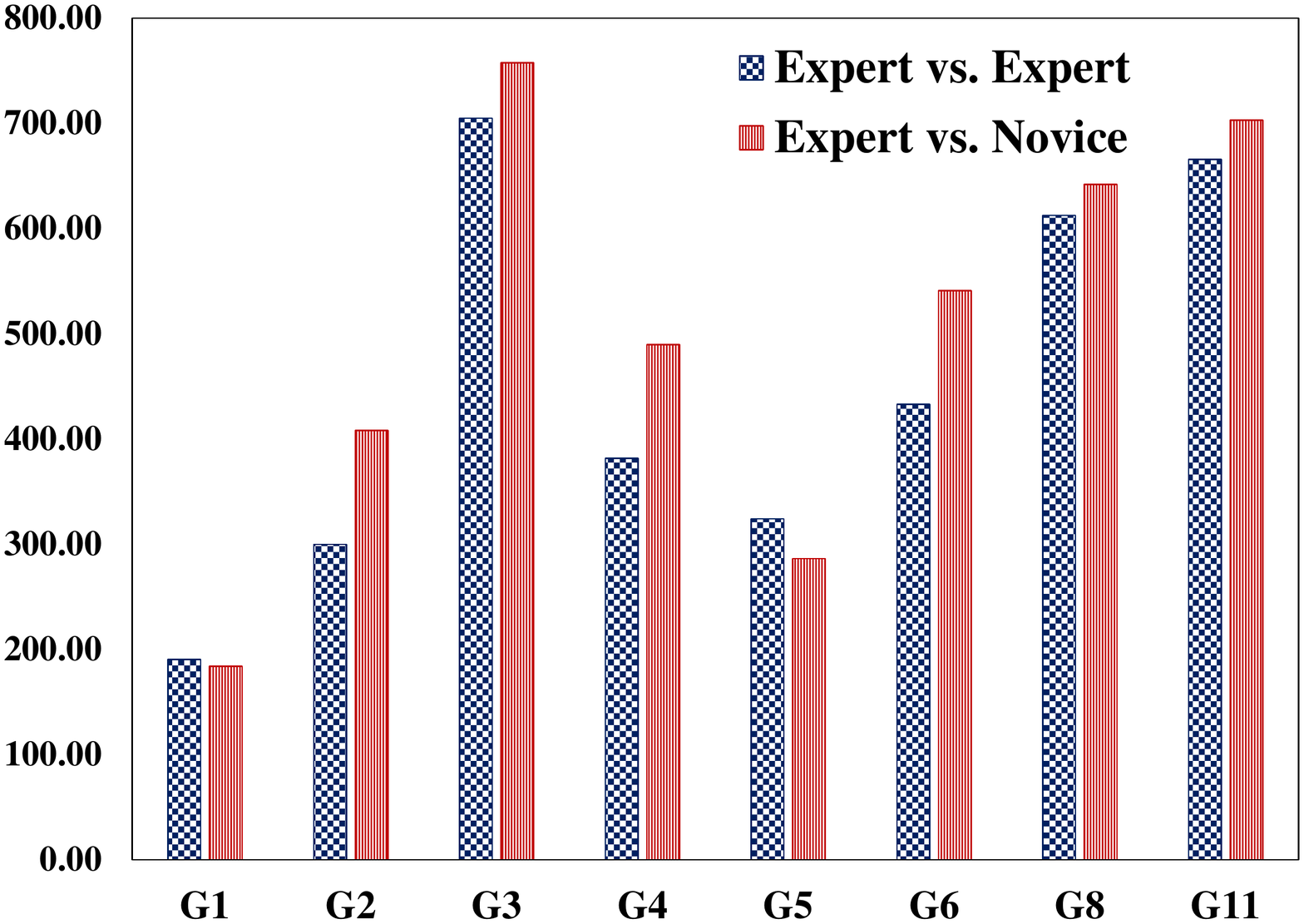}}}
\subfloat[Needle Passing]{\scalebox{.23}{\includegraphics[trim={1cm 2cm 1.2cm 2cm},clip]{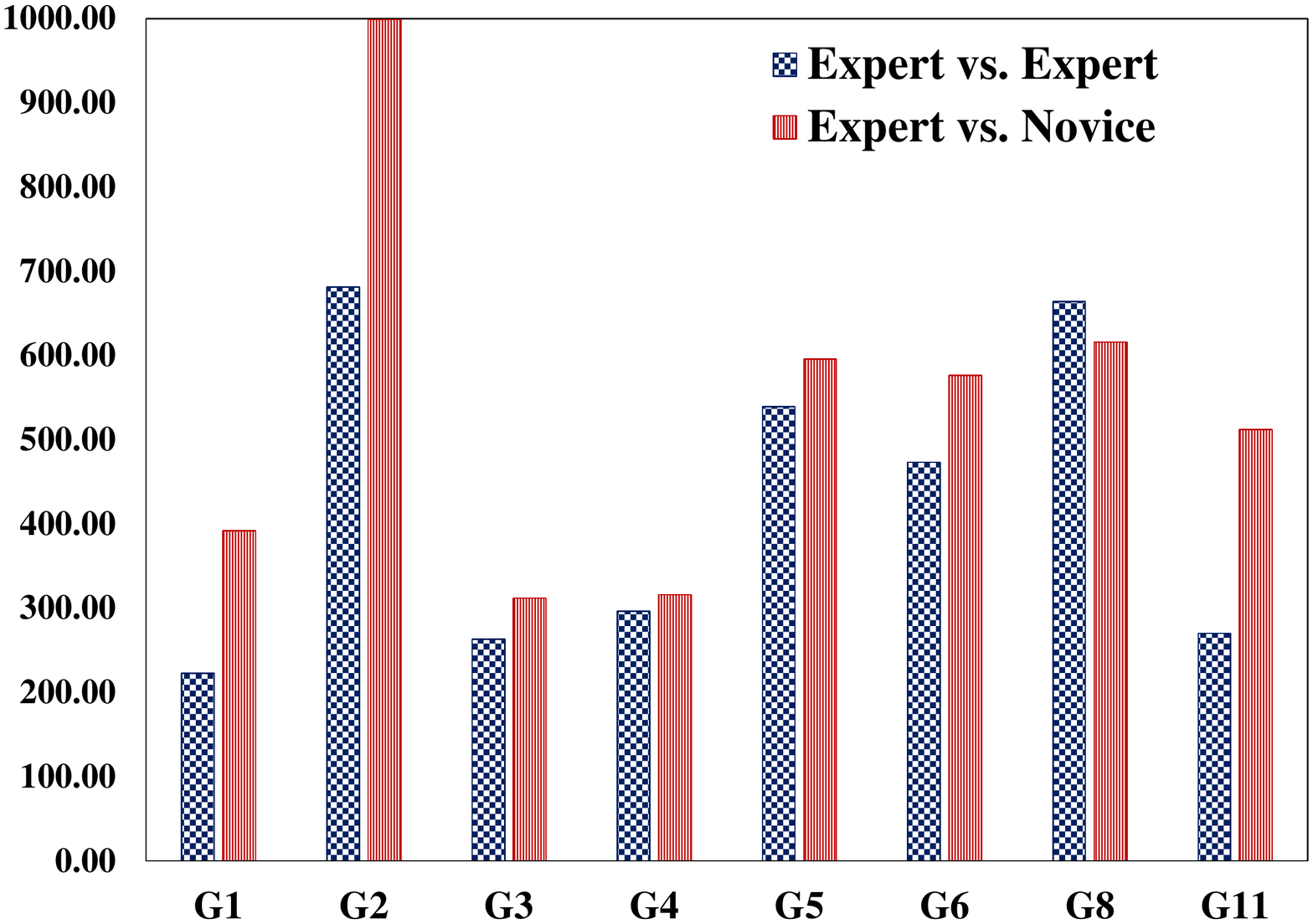}}}
\subfloat[Knot Tying]{\scalebox{.23}{\includegraphics[trim={.8cm 2cm 1.2cm 2cm},clip]{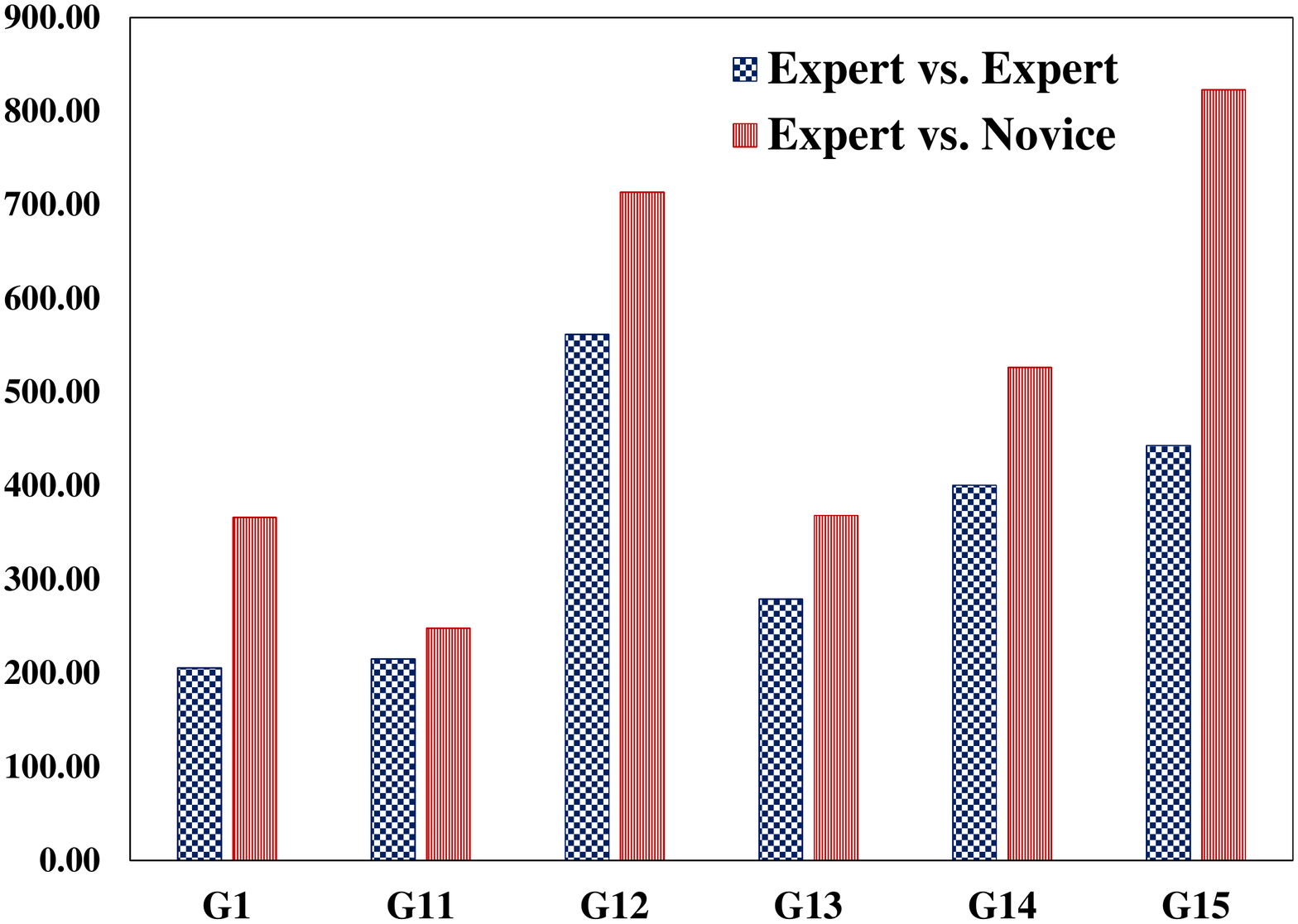}}}
\vspace{0.2em}
\caption{Average DTW distance for each surgical gesture within Experts surgeons and Experts versus  Novices for three RMIS tasks.}
\label{dist-all}%
\end{figure*}

In order to have a better understanding about experts' and novices' pattern during RMIS tasks, we compute the pairwise DTW distance within group of expert surgeons and compare it with DTW distance between novices and experts. Figure \ref{fig:boxplot} presents the boxplot for different RMIS tasks. It shows that expert surgeons do the tasks in a more similar pattern compare to novices. This conclusion is also valid for each surgical gestures separately as it shows in Figure \ref{dist-all}. It also indicates the feasibility of using DTW distance measure as a skill evaluation metric. These results align with intuition behind the DTW distance approach that proposed in this paper for gesture classification and skill assessment.

\subsection{Surgical Gesture Classification}
Table \ref{result2} shows the accuracy of the proposed method obtained for each surgeon doing different tasks. 
From the Table, it can be observed that knot tying has the minimum overall standard deviation which suggests that this task is possibly perform in more similar pattern between surgeons. On the other hand, such a difference among surgeons for needle passing and suturing suggests that the experts and novices can be separated more distinctly than knot tying.

\begin{table}[!htbp]
\centering
\caption{Accuracy of proposed method for different surgeons along with standard deviation.}
\renewcommand{\arraystretch}{1.2}
\begin{tabular}{cccc}
\cline{2-4} 
 & \textbf{Suturing} & \textbf{Needle Passing} & \textbf{Knot Tying}\tabularnewline
\hline 
\hline 
\textbf{S1} & 82.82 & 66.95 & 78.24 \tabularnewline
\textbf{S2} & 83.72 & 73.86 & 82.79 \tabularnewline
\textbf{S3} & 68.52 & 63.98 & 83.10 \tabularnewline
\textbf{S4} & 74.37 & 53.52 & 95.84 \tabularnewline
\textbf{S5} & 88.93 & 75.94 & 83.12 \tabularnewline
\textbf{S6} & 77.05 & -     & 81.69 \tabularnewline
\textbf{S7} & 86.69 & 79.90 & 89.14 \tabularnewline
\textbf{S8} & 84.78 & 77.73 & 79.17 \tabularnewline
\hline 
\textbf{Avg.} & \textbf{80.49} & \textbf{70.12} & \textbf{85.14}\tabularnewline
\hline
\textbf{Std.} & \textbf{6.47} & \textbf{8.65} & \textbf{5.35}\tabularnewline
\hline 
\end{tabular}
\label{result2}
\end{table}

\subsection{Personalized Skill Assessment}
Finally, we examine the proposed personalized skill assessment framework for RMIS tasks. First, the result in Table \ref{result2} should be expanded for each surgical gesture. Thus, we need to tabulated gesture classifications outcomes into a corresponding confusion matrix. Tables \ref{conf-su}-\ref{conf-kt} show the confusion matrix for these three tasks. The diagonal shows the correctly classified surgical gestures. Results in Table \ref{conf-su} and \ref{conf-np} show that for suturing and needle passing, gestures G5 and G8 have both low recall and precision. This suggests that these two surgical gestures might need more training or on the other hand, they might not be defined properly. 
It is good to mention that G1 and G11 can be considered as an idle position for both suturing and and needle passing.
For knot tying, Table \ref{conf-kt} suggests more training for gesture G12 and more G1 and G11 for more precise definition for this task.

\begin{table}[!htbp]
\centering
\caption{Confusion Matrix for Suturing when the experiment was performed on multiple surgeons (LOUO).}
\renewcommand{\arraystretch}{2}
\resizebox{\columnwidth}{!}{
\begin{tabular}{|c|cccccccc|c|}
\hline 
 & \textbf{\large{}G1} & \textbf{\large{}G2} & \textbf{\large{}G3} & \textbf{\large{}G4} & \textbf{\large{}G5} & \textbf{\large{}G6} & \textbf{\large{}G8} & \textbf{\large{}G11} & \textbf{\large{}Recall}\tabularnewline
\hline 
\textbf{\large{}G1} & {\large{}1240} & {\large{}75} & {\large{}0} & {\large{}0} & {\large{}0} & {\large{}0} & {\large{}0} & {\large{}0} & \textbf{\large{}0.94}\tabularnewline
\textbf{\large{}G2} & {\large{}571} & {\large{}6452} & {\large{}130} & {\large{}2} & {\large{}94} & {\large{}0} & {\large{}0} & {\large{}0} & \textbf{\large{}0.89}\tabularnewline
\textbf{\large{}G3} & {\large{}37} & {\large{}596} & {\large{}5775} & {\large{}11} & {\large{}1038} & {\large{}17} & {\large{}66} & {\large{}34} & \textbf{\large{}0.76}\tabularnewline
\textbf{\large{}G4} & {\large{}177} & {\large{}29} & {\large{}0} & {\large{}4453} & {\large{}47} & {\large{}287} & {\large{}509} & {\large{}2} & \textbf{\large{}0.79}\tabularnewline
\textbf{\large{}G5} & {\large{}244} & {\large{}512} & {\large{}0} & {\large{}0} & {\large{}909} & {\large{}0} & {\large{}33} & {\large{}0} & \textbf{\large{}0.53}\tabularnewline
\textbf{\large{}G6} & {\large{}433} & {\large{}213} & {\large{}1} & {\large{}245} & {\large{}0} & {\large{}6577} & {\large{}54} & {\large{}3} & \textbf{\large{}0.78}\tabularnewline
\textbf{\large{}G8} & {\large{}241} & {\large{}223} & {\large{}73} & {\large{}681} & {\large{}63} & {\large{}168} & {\large{}725} & {\large{}0} & \textbf{\large{}0.33}\tabularnewline
\textbf{\large{}G11} & {\large{}237} & {\large{}238} & {\large{}67} & {\large{}172} & {\large{}25} & {\large{}32} & {\large{}25} & {\large{}999} & \textbf{\large{}0.55}\tabularnewline
\hline 
\textbf{\large{}Precision} & \textbf{\large{}0.39} & \textbf{\large{}0.77} & \textbf{\large{}0.95} & \textbf{\large{}0.80} & \textbf{\large{}0.42} & \textbf{\large{}0.93} & \textbf{\large{}0.51} & \textbf{\large{}0.96} & \tabularnewline
\hline 
\end{tabular}
}
\label{conf-su}%
\end{table}%

\begin{table}[!htbp]
\centering
\caption{Confusion Matrix for Needle Passing when the experiment was performed on multiple surgeons (LOUO).}
\renewcommand{\arraystretch}{2}
\resizebox{\columnwidth}{!}{
\begin{tabular}{|c|cccccccc|c|}
\hline 
 & \textbf{\large{}G1} & \textbf{\large{}G2} & \textbf{\large{}G3} & \textbf{\large{}G4} & \textbf{\large{}G5} & \textbf{\large{}G6} & \textbf{\large{}G8} & \textbf{\large{}G11} & \textbf{\large{}Recall}\tabularnewline
\hline 
\textbf{\large{}G1} & {\large{}1237} & {\large{}0} & {\large{}51} & {\large{}35} & {\large{}49} & {\large{}34} & {\large{}40} & {\large{}0} & \textbf{\large{}0.86}\tabularnewline
\textbf{\large{}G2} & {\large{}0} & {\large{}2116} & {\large{}2669} & {\large{}268} & {\large{}236} & {\large{}146} & {\large{}199} & {\large{}0} & \textbf{\large{}0.37}\tabularnewline
\textbf{\large{}G3} & {\large{}13} & {\large{}78} & {\large{}4695} & {\large{}295} & {\large{}4} & {\large{}197} & {\large{}3} & {\large{}0} & \textbf{\large{}0.89}\tabularnewline
\textbf{\large{}G4} & {\large{}0} & {\large{}0} & {\large{}317} & {\large{}2943} & {\large{}0} & {\large{}621} & {\large{}157} & {\large{}0} & \textbf{\large{}0.73}\tabularnewline
\textbf{\large{}G5} & {\large{}23} & {\large{}20} & {\large{}566} & {\large{}177} & {\large{}566} & {\large{}38} & {\large{}60} & {\large{}50} & \textbf{\large{}0.38}\tabularnewline
\textbf{\large{}G6} & {\large{}17} & {\large{}75} & {\large{}384} & {\large{}1458} & {\large{}24} & {\large{}3263} & {\large{}163} & {\large{}0} & \textbf{\large{}0.61}\tabularnewline
\textbf{\large{}G8} & {\large{}0} & {\large{}47} & {\large{}62} & {\large{}700} & {\large{}150} & {\large{}146} & {\large{}245} & {\large{}0} & \textbf{\large{}0.18}\tabularnewline
\textbf{\large{}G11} & {\large{}250} & {\large{}0} & {\large{}0} & {\large{}21} & {\large{}0} & {\large{}11} & {\large{}0} & {\large{}914} & \textbf{\large{}0.76}\tabularnewline
\hline 
\textbf{\large{}Precision} & \textbf{\large{}0.80} & \textbf{\large{}0.91} & \textbf{\large{}0.54} & \textbf{\large{}0.51} & \textbf{\large{}0.55} & \textbf{\large{}73.23} & \textbf{\large{}0.28} & \textbf{\large{}0.95} & \tabularnewline
\hline 
\end{tabular}
}
\label{conf-np}%
\end{table}%

\begin{table}[!htbp]
\centering
\caption{Confusion Matrix for Knot Tying when the experiment was performed on multiple surgeons (LOUO).}
\renewcommand{\arraystretch}{1.4}
\resizebox{\columnwidth}{!}{
\begin{tabular}{|c|cccccc|c|}
\hline 
 & \textbf{G1} & \textbf{G11} & \textbf{G12} & \textbf{G13} & \textbf{G14} & \textbf{G15} & \textbf{Recall}\tabularnewline
\hline 
\textbf{G1} & 674 & 0 & 53 & 93 & 130 & 0 & \textbf{0.71}\tabularnewline
\textbf{G11} & 0 & 1560 & 0 & 14 & 136 & 0 & \textbf{0.91}\tabularnewline
\textbf{G12} & 282 & 340 & 1704 & 667 & 291 & 0 & \textbf{0.52}\tabularnewline
\textbf{G13} & 3 & 155 & 270 & 2939 & 167 & 0 & \textbf{0.83}\tabularnewline
\textbf{G14} & 127 & 252 & 111 & 206 & 3927 & 61 & \textbf{0.84}\tabularnewline
\textbf{G15} & 81 & 354 & 158 & 297 & 312 & 2250 & \textbf{0.65}\tabularnewline
\hline 
\textbf{Precision} & \textbf{0.58} & \textbf{0.59} & \textbf{0.74} & \textbf{0.69} & \textbf{0.79} & \textbf{0.97} & \tabularnewline
\hline 
\end{tabular}
}
\label{conf-kt}%
\end{table}%

\begin{figure*}
\centering
\subfloat[SU-Expert]{\scalebox{.23}{\includegraphics[trim={0cm 2.5cm 0cm 0cm},clip]{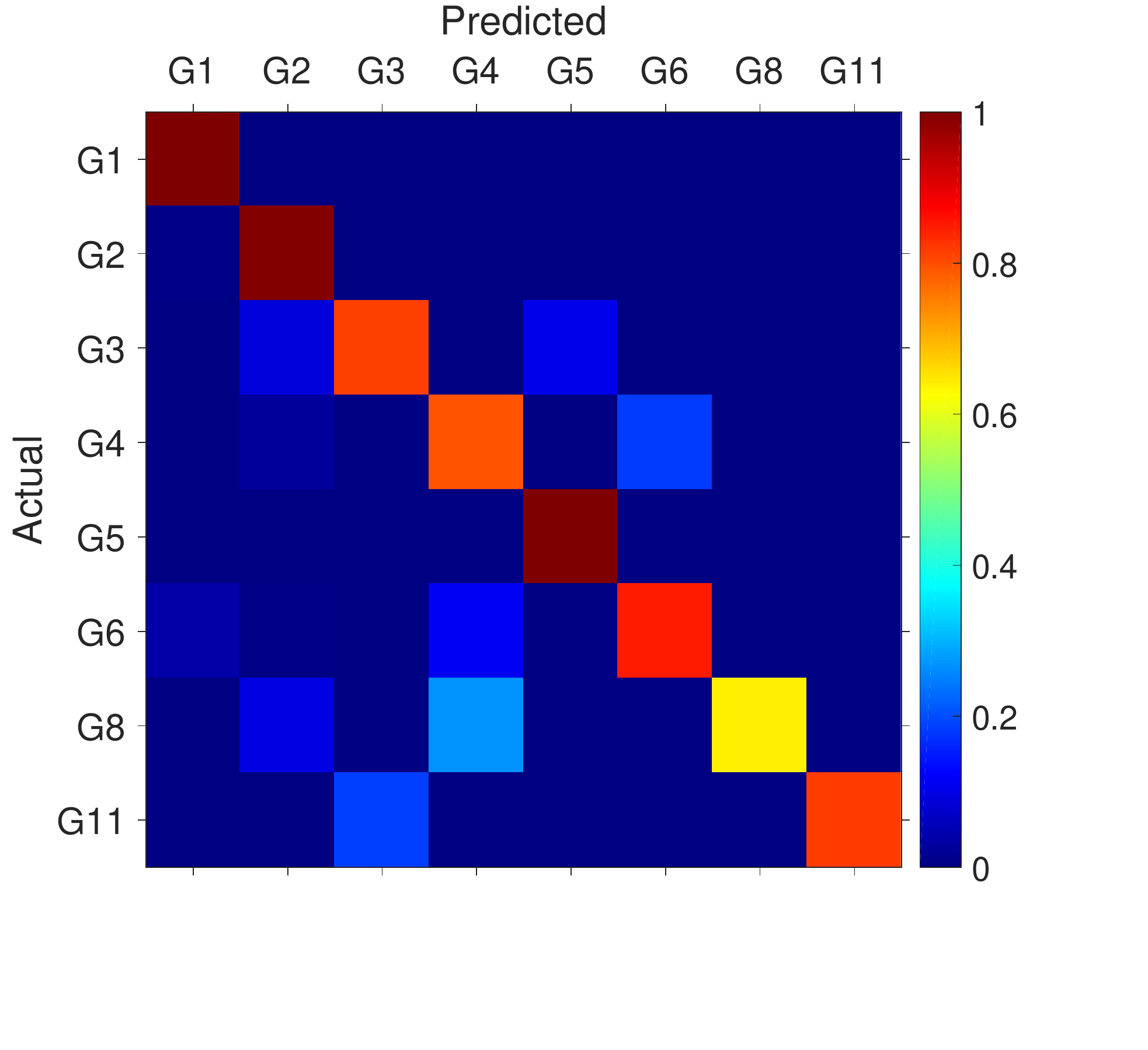}}}
\subfloat[NP-Expert]{\scalebox{.23}{\includegraphics[trim={0cm 2.5cm 0cm 0cm},clip]{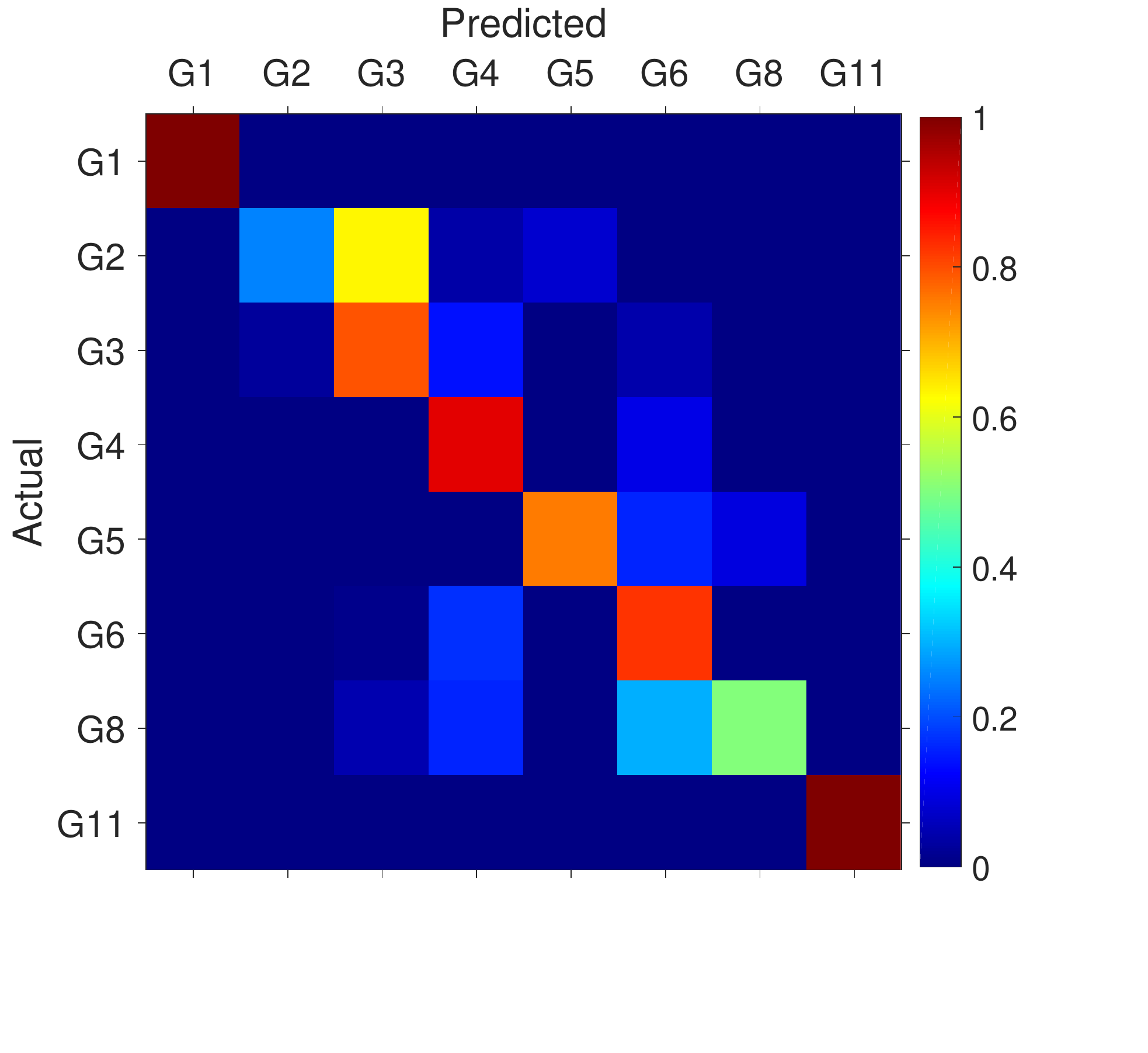}}}
\subfloat[KT-Expert]{\scalebox{.25}{\includegraphics[trim={0cm 3.3cm 0cm 0cm},clip]{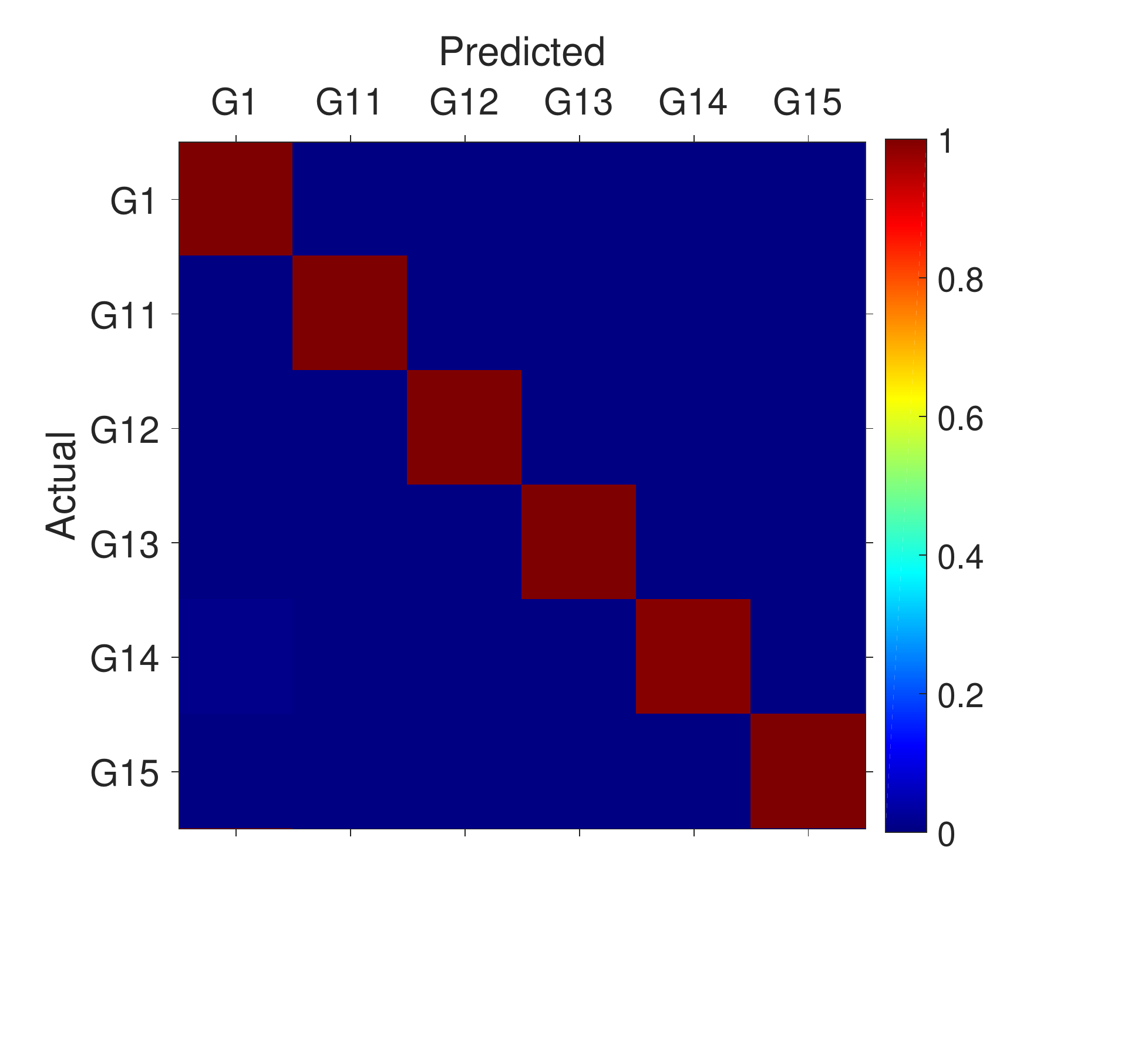}}}\\
\subfloat[SU-Novice]{\scalebox{.23}{\includegraphics[trim={0cm 2.5cm 0cm 0cm},clip]{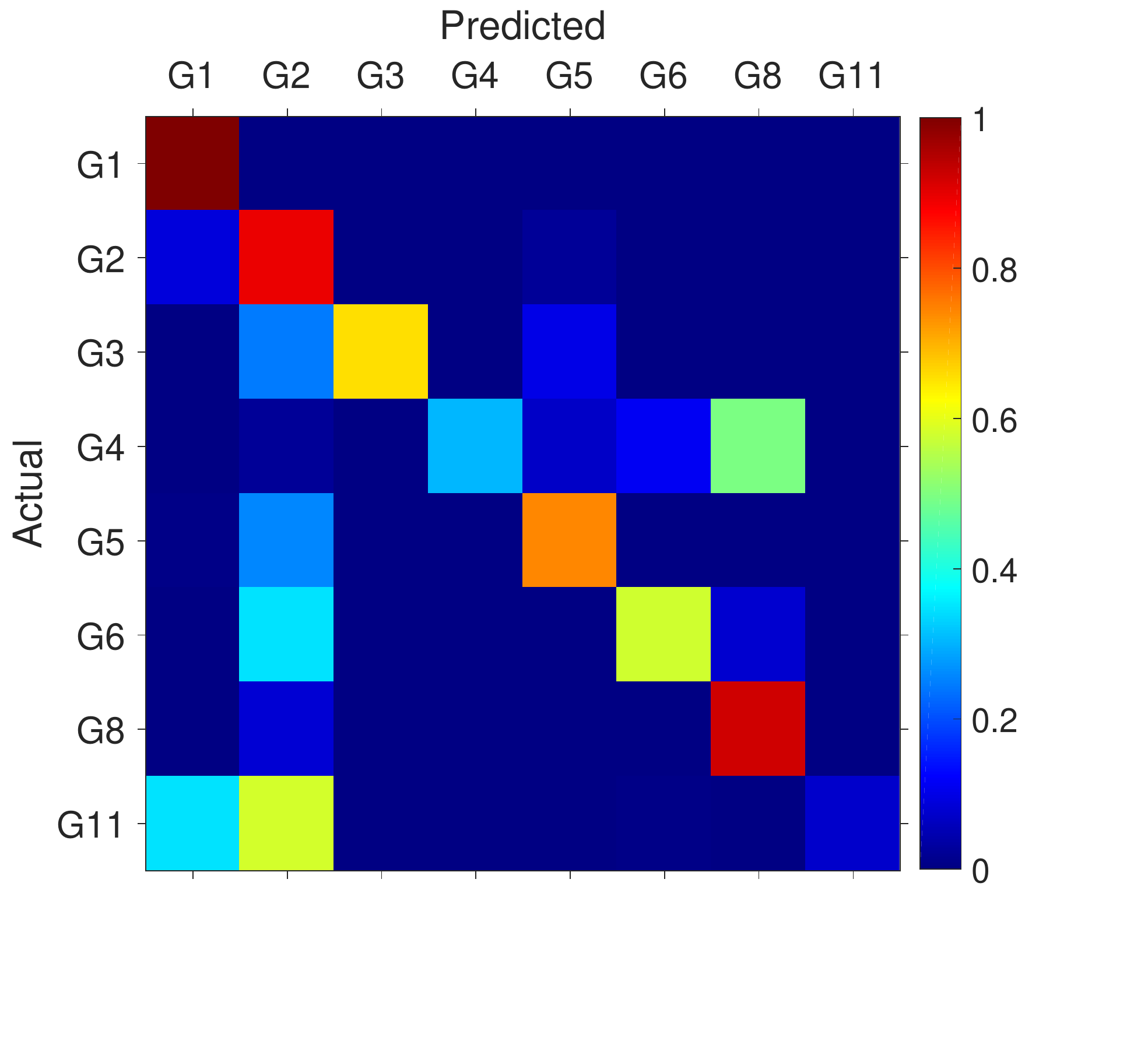}}}
\subfloat[NP-Novice]{\scalebox{.23}{\includegraphics[trim={0cm 2.5cm 0cm 0cm},clip]{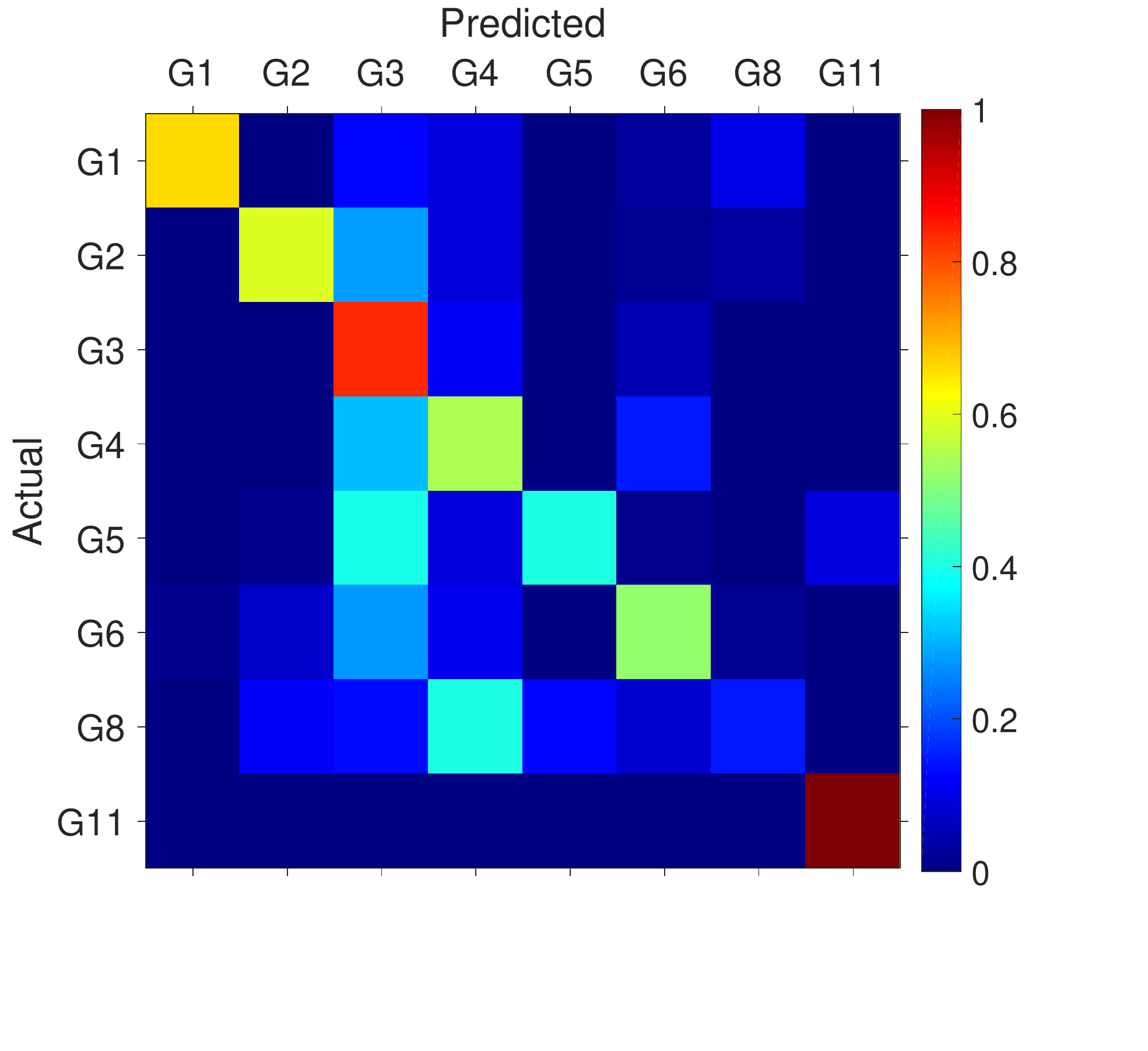}}}
\subfloat[KT-Novice]{\scalebox{.25}{\includegraphics[trim={0cm 3.3cm 0cm 0cm},clip]{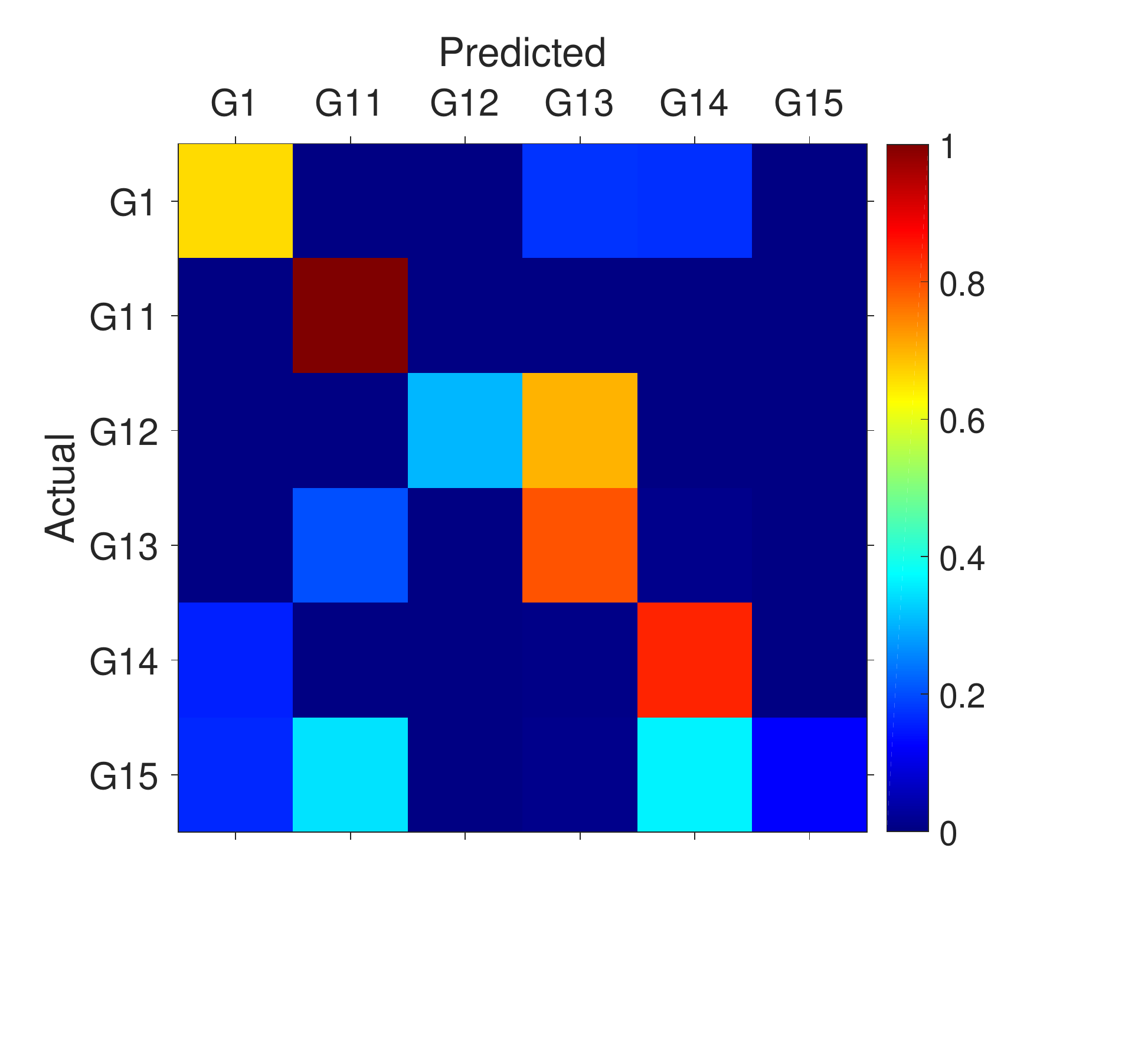}}}\\
\vspace{0.5em}
\caption{Heatmap for confusion matrix of two level of skilled surgeon doing Suturing, Needle passing and Knot tying (top row are experts and bottom row are novices).}
 \label{heat-all}%
\end{figure*}

In order to have a personalized skill assessment system that is capable of providing online feedback to surgeon during training, one should find for each surgeon, which surgical gesture does not have the same pattern as expert surgeon. In other words, which surgical gesture does not recognize correctly. In this regard, detail confusion matrix in the form of heatmap for an expert and a novice surgeon shows in Figures \ref{heat-all}. From this Figure, as an example, we can observed that compare to expert, a novice surgeon who do suturing might need more training for gesture G4, G6 and G3. One can also conclude that Needle passing is the most challenging task and the novice surgeon needs more training for almost all gesture except G3. On the other hand, Knot tying seems to be the most straight forward task and the results would suggest more training for G12 for the novice surgeon. 

\section{Conclusion}
Despite the tremendous enhancements in adequacy of surgical treatments in the recent years, the criteria of assessing surgeon's surgical skill remains subjective. With the advent of robotic surgery, the need for trained surgeons increase. Consequently, there is a huge demand for objectively evaluation of surgical skill and improve surgical training efficiency. On the other hand, the importance of recognizing gestures during surgery is self-evident by the fact that many applications deal with motion and gesture signal. In this paper, we proposed a personalized skill assessment and training framework based on time series similarity measure algorithm. We developed surgical gesture recognition model on temporal kinematic signal of robotic-assisted surgery. Based on the proposed framework, we built automated personalized RMIS gesture training system which provide online augmented feedback.  
Despite the simplicity of the proposed method, it have been shown in literature that it is difficult to beat \cite{schafer2014towards}. The performance of the proposed framework based on the experimental results are encouraging with the accuracy of approximately 80\% for suturing, 70\% for needle passing and 85\% for knot tying. These results establish the feasibility of applying time series classification methods on RMIS temporal kinematic signal data to recognize different surgical gestures during robotic minimally invasive surgery. A key advantage of our approach is its simplicity by using directly on the temporal kinematic signal of robotic-assisted surgery. However, there may be utility in extending our work by adding noise or other tasks (beside those in the training set) to the data in order to build a more robust gesture recognition method.

It should be noted that, from the results in this paper one can conclude that the global surgical gesture dictionary and their definition (Table \ref{gesture}) need modification in order to have a universal language of surgery.
More importantly, the accuracy and the robustness of any supervised classification method relies on {\em a priori} labels which are the ground truth in any machine learning methods. 
In this experiment, the labels are given by expert surgeons. This implies a subjective based annotation for surgical gesture labelling system. Though, imprecise labels result in significantly different classification accuracy. Furthermore, reliable classification is possible in light of the fact that the model learn its parameters from the precise training data with less human involvement. 
One way to overcome to this challenge is to build an unsupervised classification model to automatically decompose task to its gestures \cite{fard2017eee}. 

This paper present a step toward automatic recognition of surgical gesture also provide an insight about online feedback during training. Thus, perhaps most excitingly, the proposed framework can lay the groundwork towards development of semi-autonomous robot behaviors, such as automatic camera control during robotic-assisted surgery by online recognizing the surgical gestures that is being performed \cite{fard2016distance}.
Additionally, human factor study should be developed to have better understanding of this aspect in surgical training \cite{ellis2012management, yang2012using}.
Our intuitive approach for finding similarities between two time series queries is based on DTW distance which directly applied on the temporal kinematic signal data. Despite the promising result in this paper, a future technical challenge will be to build a more generalized models that are capable of capturing abnormal pattern of surgeon during surgery by applying rare event classification approaches.

\bibliographystyle{BibTeXtran}  
\small \bibliography{ref}  

\begin{thebibliography}{10}
\providecommand{\url}[1]{#1}
\csname url@samestyle\endcsname
\providecommand{\newblock}{\relax}
\providecommand{\bibinfo}[2]{#2}
\providecommand{\BIBentrySTDinterwordspacing}{\spaceskip=0pt\relax}
\providecommand{\BIBentryALTinterwordstretchfactor}{4}
\providecommand{\BIBentryALTinterwordspacing}{\spaceskip=\fontdimen2\font plus
\BIBentryALTinterwordstretchfactor\fontdimen3\font minus
  \fontdimen4\font\relax}
\providecommand{\BIBforeignlanguage}[2]{{%
\expandafter\ifx\csname l@#1\endcsname\relax
\typeout{** WARNING: IEEEtran.bst: No hyphenation pattern has been}%
\typeout{** loaded for the language `#1'. Using the pattern for}%
\typeout{** the default language instead.}%
\else
\language=\csname l@#1\endcsname
\fi
#2}}
\providecommand{\BIBdecl}{\relax}
\BIBdecl

\bibitem{reznick1993teaching}
R.~K. Reznick, ``Teaching and testing technical skills,'' \emph{The American
  journal of surgery}, vol. 165, no.~3, pp. 358--361, 1993.

\bibitem{darzi1999assessing}
A.~Darzi, S.~Smith, and N.~Taffinder, ``Assessing operative skill: needs to
  become more objective,'' \emph{BMJ: British Medical Journal}, vol. 318, no.
  7188, p. 887, 1999.

\bibitem{Kassahun2015}
Y.~Kassahun, B.~Yu, A.~T. Tibebu, D.~Stoyanov, S.~Giannarou, J.~H. Metzen, and
  E.~{Vander Poorten}, ``{Surgical robotics beyond enhanced dexterity
  instrumentation: a survey of machine learning techniques and their role in
  intelligent and autonomous surgical actions},'' \emph{International Journal
  of Computer Assisted Radiology and Surgery}, vol.~11, no.~4, pp. 553--568,
  2016.

\bibitem{ellis2015task}
R.~D. Ellis, A.~J. Munaco, L.~A. Reisner, M.~D. Klein, A.~M. Composto, A.~K.
  Pandya, and B.~W. King, ``Task analysis of laparoscopic camera control
  schemes,'' \emph{The International Journal of Medical Robotics and Computer
  Assisted Surgery}, 2015, DOI:10.1002/rcs.1716.

\bibitem{van2010objective}
P.~Van~Hove, G.~Tuijthof, E.~Verdaasdonk, L.~Stassen, and J.~Dankelman,
  ``Objective assessment of technical surgical skills,'' \emph{British journal
  of surgery}, vol.~97, no.~7, pp. 972--987, 2010.

\bibitem{lalys2014surgical}
F.~Lalys and P.~Jannin, ``Surgical process modelling: a review,''
  \emph{International journal of computer assisted radiology and surgery},
  vol.~9, no.~3, pp. 495--511, 2014.

\bibitem{guthart2000intuitivetm}
G.~Guthart and J.~K. Salisbury~Jr, ``The intuitivetm telesurgery system:
  Overview and application.'' in \emph{ICRA}, 2000, pp. 618--621.

\bibitem{fard2017eee}
M.~J. Fard, S.~Ameri, R.~B. Chinnam, and R.~D. Ellis, ``Soft boundary approach
  for unsupervised gesture segmentation in robotic-assisted surgery,''
  \emph{IEEE Robotics and Automation Letters}, vol.~2, no.~1, pp. 171--178, Jan
  2017.

\bibitem{jahanbani2016computational}
M.~Jahanbani~Fard, ``Computational modeling approaches for task analysis in
  robotic-assisted surgery,'' 2016.

\bibitem{reiley2011review}
C.~E. Reiley, H.~C. Lin, D.~D. Yuh, and G.~D. Hager, ``Review of methods for
  objective surgical skill evaluation,'' \emph{Surgical endoscopy}, vol.~25,
  no.~2, pp. 356--366, 2011.

\bibitem{rosen2001markov}
J.~Rosen, B.~Hannaford, C.~G. Richards, and M.~N. Sinanan, ``Markov modeling of
  minimally invasive surgery based on tool/tissue interaction and force/torque
  signatures for evaluating surgical skills,'' \emph{Biomedical Engineering,
  IEEE Transactions on}, vol.~48, no.~5, pp. 579--591, 2001.

\bibitem{reiley2009task}
C.~E. Reiley and G.~D. Hager, ``Task versus subtask surgical skill evaluation
  of robotic minimally invasive surgery,'' in \emph{Medical Image Computing and
  Computer-Assisted Intervention--MICCAI 2009}.\hskip 1em plus 0.5em minus
  0.4em\relax Springer, 2009, pp. 435--442.

\bibitem{Zappella2013}
L.~Zappella, B.~B\'{e}jar, G.~Hager, and R.~Vidal, ``{Surgical gesture
  classification from video and kinematic data.}'' \emph{Medical image
  analysis}, vol.~17, no.~7, pp. 732--45, Oct. 2013.

\bibitem{Ahmidi2015}
N.~Ahmidi, P.~Poddar, J.~D. Jones, S.~S. Vedula, L.~Ishii, G.~D. Hager, and
  M.~Ishii, ``{Automated objective surgical skill assessment in the operating
  room from unstructured tool motion in septoplasty},'' \emph{International
  Journal of Computer Assisted Radiology and Surgery}, vol.~10, no.~6, pp.
  981--991, 2015.

\bibitem{schout2010validation}
B.~Schout, A.~Hendrikx, F.~Scheele, B.~Bemelmans, and A.~Scherpbier,
  ``Validation and implementation of surgical simulators: a critical review of
  present, past, and future,'' \emph{Surgical endoscopy}, vol.~24, no.~3, pp.
  536--546, 2010.

\bibitem{fu2011review}
T.-c. Fu, ``A review on time series data mining,'' \emph{Engineering
  Applications of Artificial Intelligence}, vol.~24, no.~1, pp. 164--181, 2011.

\bibitem{RCS:RCS1766}
\BIBentryALTinterwordspacing
M.~J. Fard, A.~K. Pandya, R.~B. Chinnam, M.~D. Klein, and R.~D. Ellis,
  ``Distance-based time series classification approach for task recognition
  with application in surgical robot autonomy,'' \emph{The International
  Journal of Medical Robotics and Computer Assisted Surgery}, pp. n/a--n/a,
  2016, rCS-16-0026.R2. [Online]. Available:
  \url{http://dx.doi.org/10.1002/rcs.1766}
\BIBentrySTDinterwordspacing

\bibitem{fard2016machine}
M.~J. Fard, S.~Ameri, A.~K. Chinnam, Ratna B.~Pandya, M.~D. Klein, and R.~D.
  Ellis, ``Machine learning approach for skill evaluation in robotic-assisted
  surgery,'' in \emph{Lecture Notes in Engineering and Computer Science:
  Proceedings of The World Congress on Engineering and Computer Science 2016},
  2016, pp. 433--437.

\bibitem{bernad1996finding}
D.~Bernad, ``Finding patterns in time series: a dynamic programming approach,''
  \emph{Advances in knowledge discovery and data mining}, 1996.

\bibitem{bhatia2010survey}
N.~Bhatia \emph{et~al.}, ``Survey of nearest neighbor techniques,''
  \emph{International Journal of Computer Science and Information Security},
  vol.~8, no.~2, pp. 302--305, 2010.

\bibitem{mahtabIEOM2015}
M.~J. Fard, S.~Ameri, and A.~Zeinal~Hamadani, ``Bayesian approach for early
  stage reliability prediction of evolutionary products,'' in \emph{Proceedings
  of the International Conference on Operations Excellence and Service
  Engineering}.\hskip 1em plus 0.5em minus 0.4em\relax Orlando, Florida, USA,
  2015, pp. 361--371.

\bibitem{fard2016early}
M.~J. Fard, S.~Chawla, and C.~K. Reddy, ``Early-stage event prediction for
  longitudinal data,'' in \emph{Pacific-Asia Conference on Knowledge Discovery
  and Data Mining}.\hskip 1em plus 0.5em minus 0.4em\relax Springer, 2016, pp.
  139--151.

\bibitem{7564399}
M.~J. Fard, P.~Wang, S.~Chawla, and C.~K. Reddy, ``A bayesian perspective on
  early stage event prediction in longitudinal data,'' \emph{IEEE Transactions
  on Knowledge and Data Engineering}, vol.~28, no.~12, pp. 3126--3139, Dec
  2016.

\bibitem{keogh2003need}
E.~Keogh and S.~Kasetty, ``On the need for time series data mining benchmarks:
  a survey and empirical demonstration,'' \emph{Data Mining and knowledge
  discovery}, vol.~7, no.~4, pp. 349--371, 2003.

\bibitem{lin2012pattern}
J.~Lin, S.~Williamson, K.~Borne, and D.~DeBarr, ``Pattern recognition in time
  series,'' \emph{Advances in Machine Learning and Data Mining for Astronomy},
  vol.~1, pp. 617--645, 2012.

\bibitem{Chaovalitwongse2007}
W.~A. Chaovalitwongse, Y.-j. Fan, S.~Member, and R.~C. Sachdeo, ``{On the Time
  Series K -Nearest Neighbor Classification of Abnormal Brain Activity},''
  \emph{IEEE Transactions on Systems, Man and Cybernetics, Part A: System and
  Humans}, vol.~37, no.~6, pp. 1005--1016, 2007.

\bibitem{gaojhu}
Y.~Gao, S.~S. Vedula, C.~E. Reiley, N.~Ahmidi, B.~Varadarajan, H.~C. Lin,
  L.~Tao, L.~Zappella, B.~B{\'e}jar, D.~D. Yuh \emph{et~al.}, ``{JHU-ISI}
  gesture and skill assessment working set ({JIGSAWS}): A surgical activity
  dataset for human motion modeling,'' in \emph{Modeling and Monitoring of
  Computer Assisted Interventions ({M2CAI})– {MICCAI} Workshop}, 2014.

\bibitem{Reiley2008}
C.~E. Reiley, H.~C. Lin, B.~Varadarajan, B.~Vagvolgyi, S.~Khudanpur, D.~D. Yuh,
  and G.~D. Hager, ``{Automatic recognition of surgical motions using
  statistical modeling for capturing variability},'' \emph{Studies in health
  technology and informatics}, vol. 132, no.~1, pp. 396--401, Jan. 2008.

\bibitem{schafer2014towards}
P.~Sch{\"a}fer, ``Towards time series classification without human
  preprocessing,'' in \emph{Machine Learning and Data Mining in Pattern
  Recognition}.\hskip 1em plus 0.5em minus 0.4em\relax Springer, 2014, pp.
  228--242.

\bibitem{ellis2012management}
R.~D. Ellis, M.~J. Fard, K.~Yang, W.~Jordan, N.~Lightner, and S.~Yee,
  ``Management of medical equipment reprocessing procedures: A human
  factors/system reliability perspective,'' in \emph{Advances in Human Aspects
  of Healthcare}.\hskip 1em plus 0.5em minus 0.4em\relax CRC Press, 2012, pp.
  689--698.

\bibitem{yang2012using}
K.~Yang, N.~Lightner, S.~Yee, M.~J. Fard, and W.~Jordan, ``Using computerized
  technician competency validation to improve reusable medical equipment
  reprocessing system reliability,'' in \emph{Advances in Human Aspects of
  Healthcare}.\hskip 1em plus 0.5em minus 0.4em\relax CRC Press, 2012, pp.
  556--564.

\end{thebibliography}

\end{document}